\documentclass[letterpaper]{article} 
\usepackage{aaai2026}  
\usepackage{times}  
\usepackage{helvet}  
\usepackage{courier}  
\usepackage[hyphens]{url}  
\usepackage{graphicx} 
\usepackage{natbib}  
\usepackage{caption} 
\usepackage{algorithm}
\usepackage{algorithmic}

\usepackage{amsmath}
\usepackage{amssymb}
\usepackage{xcolor}
\usepackage{enumitem}
\usepackage{booktabs}
\usepackage{multirow}
\usepackage{caption}
\usepackage{subcaption}
\usepackage{tcolorbox}
\usepackage{tabularx}
\usepackage{makecell}

\usepackage{newfloat}
\usepackage{listings}
\DeclareCaptionStyle{ruled}{labelfont=normalfont,labelsep=colon,strut=off} 
\floatstyle{ruled}
\newfloat{listing}{tb}{lst}{}
\floatname{listing}{Listing}
\title{PathMind: A Retrieve-Prioritize-Reason Framework for Knowledge Graph Reasoning with Large Language Models}
\author{
    Yu Liu\textsuperscript{\rm 1, \rm 2}, 
    Xixun Lin\textsuperscript{\rm 1, \rm 2}\thanks{Corresponding author.},
    Yanmin Shang\textsuperscript{\rm 1, \rm 2},
    Yangxi Li\textsuperscript{\rm 3},
    Shi Wang\textsuperscript{\rm 4},
    Yanan Cao\textsuperscript{\rm 1, \rm 2}
}
\affiliations{
    \textsuperscript{\rm 1}Institute of Information Engineering, Chinese Academy of Sciences \\
    \textsuperscript{\rm 2}School of Cyber Security, University of Chinese Academy of Sciences \\
    \textsuperscript{\rm 3}Peking University, \textsuperscript{\rm 4}Institute of Computing Technology, Chinese Academy of Sciences \\
    \{liuyu2022, linxixun, shangyanmin, caoyanan\}@iie.ac.cn, liyangxi@outlook.com, wangshi@ict.ac.cn
}

\usepackage{bibentry}

\begin{document}

\maketitle

\begin{abstract}
Knowledge graph reasoning (KGR) is the task of inferring new knowledge by performing logical deductions on knowledge graphs.  Recently, large language models (LLMs) have demonstrated remarkable performance in complex reasoning tasks. Despite promising success, current LLM-based KGR methods still face two critical limitations. First, existing methods often extract reasoning paths indiscriminately, without assessing their different importance, which may introduce irrelevant noise that misleads LLMs. Second, while some methods leverage LLMs to dynamically explore potential reasoning paths, they require high retrieval demands and frequent LLM calls. To address these limitations, we propose PathMind, a novel framework designed to enhance faithful and interpretable reasoning by selectively guiding LLMs with important reasoning paths. Specifically, PathMind follows a "Retrieve-Prioritize-Reason" paradigm. First, it retrieves a query subgraph from KG through the retrieval module. Next, it introduces a path prioritization mechanism that identifies important reasoning paths using a semantic-aware path priority function, which simultaneously considers the accumulative cost and the estimated future cost for reaching the target. Finally, PathMind generates accurate and logically consistent responses via a dual-phase training strategy, including task-specific instruction tuning and path-wise preference alignment. Extensive experiments on benchmark datasets demonstrate that PathMind consistently outperforms competitive baselines, particularly on complex reasoning tasks with fewer input tokens, by identifying essential reasoning paths.
\end{abstract}


\section{Introduction}

Reasoning, the ability to derive logical conclusions from available knowledge, has been a long-standing goal in the pursuit of artificial intelligence~\citep{halpern1986reasoning}. Knowledge graphs (KGs) represent structured relationships between entities and serve as a fundamental foundation for reasoning~\citep{kim2023kg, pan2024unifying}. Knowledge graph reasoning (KGR) aims to infer new knowledge or answer complex queries based on KGs, facilitating various practical applications such as recommendation systems~\citep{zhou2020interactive}, question answering~\citep{liu2024generative}, and biomedical inference~\citep{jiang2025reasoning}. Despite great achievements, KGR remains a challenging task, as real-world KGs are often large-scale and inherently incomplete, making reasoning unreliable~\citep{ pan2024unifying}.

\begin{figure}
    \centering
    \includegraphics[width=0.47\textwidth]{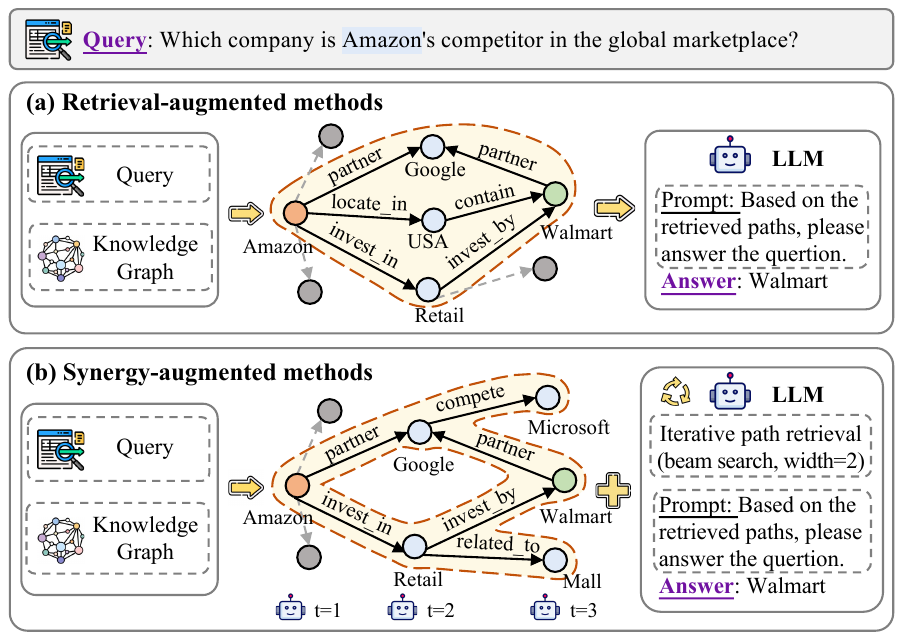}
    \caption{Illustration of LLM-based KGR methods.  (a) Retrieval-augmented methods retrieve relevant information from KGs for LLMs. (b) Synergy-augmented methods integrate KGs and LLMs through iterative interaction.} \label{fig_example}
\end{figure}

Recently, large language models (LLMs) have demonstrated remarkable performance on complex reasoning tasks~\citep{meyer2023chatgpt}, representing a significant advancement toward artificial general intelligence.  
These models leverage advanced pre-training techniques and vast amounts of unlabeled text data to understand and generate human language with impressive fluency and coherence.
Consequently, LLM-based KGR has attracted increasing interest, aiming to leverage the powerful generalizability of LLMs to perform knowledge reasoning over KGs~\citep{jiang2023structgpt, sun2024think, ao2025lightprof}.

Existing LLM-based KGR methods can be broadly divided into two main lines: \textit{retrieval-augmented} and \textit{synergy-augmented} paradigms, as illustrated in Figure~\ref{fig_example}(a) and (b). Retrieval-augmented methods~\citep{kim2023kg, luo2024reasoning, long2025eperm} retrieve query-relevant triples or multi-hop paths from KGs, which are then verbalized into LLMs to improve reasoning abilities. In contrast, synergy-augmented methods~\citep{sun2024think, chen2024plan} treat LLMs as agents that iteratively interact with KGs to explore possible reasoning paths and generate answers. In general, these two methods effectively leverage the complementary strengths of LLMs and KGs, achieving superior performance compared to previous strong approaches.

Despite their success, existing methods still suffer from several critical limitations. First, retrieval-augmented methods often extract potential reasoning paths indiscriminately, failing to evaluate their importance for answer generation, which may introduce irrelevant or noisy information that misleads LLMs. For instance, when querying \textit{"Which company is Amazon's competitor in the global marketplace?"} in Figure~\ref{fig_example}, the path {\small$Amazon \xrightarrow{invest\_in}Retail\xrightarrow{invest\_by} Walmart$} clearly indicates a competitive relationship between Amazon and Walmart, whereas less relevant paths like {\small$Amazon\xrightarrow{partner}Google\xleftarrow{partner} Walmart$} might misleadingly imply collaboration. Second, although synergy-augmented methods leverage LLMs to dynamically discover reasoning paths, they often encounter significant computational challenges, such as substantial retrieval demands in large search spaces and high overhead from multiple LLM calls, which severely limit their scalability and practicality in real-world scenarios. 

To overcome these limitations, we propose \textbf{PathMind}, a novel framework that enhances faithful and interpretable reasoning by selectively guiding LLMs with important reasoning paths. Specifically, PathMind follows a ”Retrieve-Prioritize-Reason” paradigm. Given a question, we first retrieve the query subgraph from the KG and encode it into graph representations. Next, we introduce a path prioritization mechanism that identifies important reasoning paths using a semantic-aware path priority function, which simultaneously considers the accumulative cost up to the current node and the estimated future cost to reach the target entity. Finally, we leverage these retrieved paths to guide LLMs via a dual-phase training strategy: task-specific instruction tuning and path-wise preference alignment, enabling LLMs to generate accurate and logically consistent responses without incurring the overhead of multiple LLM calls. Extensive experiments and comprehensive analyses on benchmark datasets show the superiority of PathMind.

In summary, our contributions are as follows:
\begin{itemize}
\setlength\itemsep{0em}
\item We introduce a new framework named PathMind, which leverages important reasoning paths to effectively guide LLMs toward accurate logical reasoning, enhancing both faithfulness and interpretability.
\item We propose an effective path prioritization mechanism that simultaneously models the accumulative cost and estimates future cost to identify important reasoning paths, and further improve LLMs via task-specific instruction tuning and path-wise preference alignment.
\item Experimental results on widely-used KGR benchmarks demonstrate that PathMind achieves competitive performance compared to strong baselines, particularly on complex reasoning tasks with fewer input tokens\footnote{Our code is available at \url{https://github.com/liuyudiy/PathMind}.}.
\end{itemize}

\section{Related work}

\begin{figure*}
    \centering
    \includegraphics[width=1\textwidth]{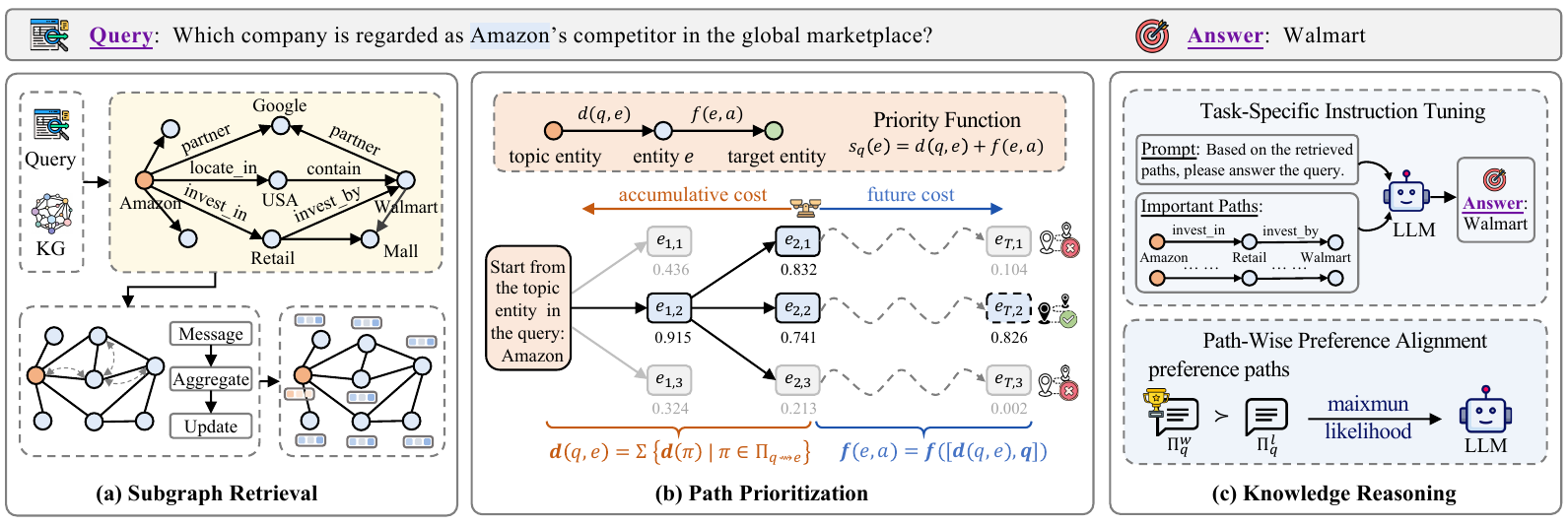}
	\caption{The overall framework of our \textbf{PathMind}. (a) Subgraph Retrieval extracts the query subgraph from the KG and encodes it into graph representations. (b) Path Prioritization identifies important reasoning paths according to a path priority function. (c) Knowledge Reasoning enhances LLM reasoning via task-specific instruction tuning and path-wise preference alignment.} \label{fig_framework}
\end{figure*}

\subsubsection{Retrieval-Augmented Methods.} These methods typically retrieve factual knowledge from KGs and integrate it into LLMs to improve reasoning abilities. Some methods~\citep{kim2023kg, li2025simple} retrieve relevant facts by evaluating semantic similarities between the question and associated triples. Other approaches~\citep{luo2024reasoning, long2025eperm, liu2025ontology} extract explicit paths to exploit structures within KGs.  In particular, RoG~\citep{luo2024reasoning} proposes a planning-retrieval-reasoning framework to generate relational paths. GNN-RAG~\citep{mavromatis2025gnn} retrieves the shortest paths between topic entities and answer candidates. GCR~\citep{luo2025graph} generates reliable reasoning paths grounded in KGs. EPERM~\citep{long2025eperm} and ORT~\citep{liu2025ontology} discover more reasoning chains to support inference.  Meanwhile, certain approaches~\citep{liu2025enhancing, ao2025lightprof} employ graph neural networks (GNNs) to learn graph embeddings for retrieved subgraphs. Overall, these methods depend heavily on the information retrieved from KGs. However, current approaches often treat them equally, failing to distinguish the varying contributions of reasoning paths, which can result in extraneous or noisy information that may mislead LLMs.

\subsubsection{Synergy-Augmented Methods.} These methods combine LLMs and KGs via an interaction mechanism that iteratively explores KGs to discover potential reasoning paths. Early approaches, such as ChatKBQA~\citep{luo2023chatkbqa} and FlexKBQA~\citep{li2024flexkbqa}, utilize LLMs to convert natural language questions into structural queries (e.g., SPARQL), which are then executed on KGs. More recently, most works~\citep{sun2024think, ma2025think}  treat LLMs as agents~\cite{lin2025llm} that interact with KGs to derive reasoning paths and generate target responses.  For example, ToG~\citep{sun2024think} employs LLMs to iteratively perform beam search on KGs; its successor, ToG-2~\citep{ma2025think}, extends this by incorporating external unstructured knowledge. PoG~\citep{chen2024plan} proposes a self-correcting adaptive planning paradigm. Additionally, techniques such as KD-CoT~\citep{wang2023knowledge} and SymAgent~\citep{liu2025symagent} explore chain-of-thought prompting to elicit deep reasoning of LLMs.  Despite their success, these methods still face significant computational challenges, stemming from vast search spaces and multiple LLM API calls, which seriously limit their practical applications.

\section{Preliminary}

\subsubsection{Knowledge Graphs.}
KGs represent factual knowledge with structured relationships. Formally, a KG can be defined as $\mathcal{G}=(\mathcal{E}, \mathcal{R},\mathcal{T})$, where $\mathcal{E}$ and $\mathcal{R}$ are the sets of entities and relations, respectively, and $\mathcal{T}=\{(e,r,e') \mid e,e' \in \mathcal{E}, r \in \mathcal{R} \}$ represents the set of relational triples. Given a query $q$ and a KG $\mathcal{G}$, the KGR task is to design a function $f$ to predict answers $a \in \mathcal{A}_q$ based on knowledge from $\mathcal{G}$, i.e., $a=f(q,\mathcal{G})$. Following previous work~\citep{luo2024reasoning}, we assume the topic entity $e_q \in \mathcal{O}_q $ observed in $q$ and the answer $a \in \mathcal{A}_q $ are labeled and linked to the corresponding entities in $\mathcal{G}$, i.e., $\mathcal{O}_q, \mathcal{A}_q \subseteq \mathcal{E}$.

\subsubsection{Reasoning Paths.} Reasoning paths are sequences of consecutive triples in KGs. More concretely, a reasoning path can be defined as $\pi =e_0\xrightarrow{r_1} e_1 \xrightarrow{r_2} \dots \xrightarrow{r_l} e_l $, where $\forall\left(e_{i-1}, r_i, e_i\right) \in \mathcal{T}$. These paths reveal the connections among entities that can support knowledge reasoning. However, in real-world KGs, the number of possible paths is extremely large, and only a small set of them is important for reasoning. Based on this observation, we introduce the concept of \textit{important reasoning paths}—those that are particularly necessary for reasoning. For example, in Figure~\ref{fig_example}, the path {\small $Amazon \xrightarrow{invest\_in} Retail \xrightarrow{invest\_by} Walmart$} suggests that Amazon and Walmart are in a competitive relationship. In contrast, the path {\small $Amazon \xrightarrow{partner} Google \xleftarrow{partner} Walmart$} provides no clear evidence of competition and may even misleadingly imply collaboration.

\subsubsection{Task Formulation.} Typically, the KGR task can be formulated as an optimization problem that maximizes the probability of inferring the answer from the KG $\mathcal{G}$ given the query $q$. By retrieving reasoning paths $\Pi$ as supporting evidence, we empower LLMs to perform structured reasoning. Formally, the task can be defined as follows:
\begin{equation}
    P(a|q, \mathcal{G}) = \sum_{\pi \in \Pi} P_{\phi}(a|q, \pi) P_{\varphi}( \pi |q, \mathcal{G}),
    \label{eq_1}
\end{equation}
\noindent where $P_{\varphi}(\pi |q, \mathcal{G})$ denotes the probability of discovering a reasoning path $\pi$ from the KG $\mathcal{G}$ given the query $q$ using the retriever model parameterized by $\varphi$. $P_{\phi}(a|q, \pi)$ is the probability of generating an answer $a$ using the LLM parameterized by $\phi$, given the query $q$ and the reasoning path $\pi$.

To acquire reasoning paths for reasoning, previous studies have followed \textit{retrieval-augmented} or \textit{synergy-augmented} paradigm. However, the former typically extracts reasoning paths equally, without evaluating their different importance to answer generation, while the latter is computationally intensive and leads to high costs. To address these issues, we propose a novel method that prioritizes important reasoning paths, enabling more reliable reasoning.

\section{Methodology}

In this paper, we introduce PathMind, a new framework that integrates LLMs and KGs to enhance faithful and interpretable reasoning. The overall architecture of PathMind is illustrated in Figure~\ref{fig_framework}. Specifically, our model consists of three key components: (1) Subgraph Retrieval, (2) Path Prioritization, and (3) Knowledge Reasoning. In the following, we explain these technical details.

\subsection{Subgraph Retrieval Module}\label{module_1}

Given a query $q$, the first step is to extract the relevant subgraph $\mathcal{G}_q=(\mathcal{E}_q, \mathcal{R}_q, \mathcal{T}_q)$ from the KG,  aiming to reduce the search space while preserving essential information. 

\subsubsection{Query Subgraph Extraction.} For each topic entity $e_q \in \mathcal{O}_q$ in the query $q$, we first retrieve its $k$-hop neighborhood $\mathcal{N}_k(e_q)$. We then take the union of  these neighborhoods as the set of subgraph nodes, denoted as $\mathcal{E}_q = \bigcup_{e_q \in \mathcal{O}_q} \mathcal{N}_k(e_q)$.  Next, we extract the set of edges $\mathcal{R}_q$ from $\mathcal{G}$ that connect nodes within $\mathcal{E}_q$. Using these nodes and edges, we obtain the triple set $\mathcal{T}_q=\{ (e, r, e') \mid e, e' \in \mathcal{E}_q$, $r \in \mathcal{R}_q \}$.  Finally, the query subgraph $\mathcal{G}_q=(\mathcal{E}_q, \mathcal{R}_q, \mathcal{T}_q)$ is constructed.

\subsubsection{Graph Representation Learning.} After constructing the query subgraph $\mathcal{G}_q$, we utilize a graph neural network (GNN) to learn the representations of nodes and relations. The GNN leverages message passing and aggregation mechanisms to capture complex relationships and structural dependencies inherent in $\mathcal{G}_q$. Specifically, at each layer $l$, the representation of node $e \in \mathcal{E}_q$ is updated as follows:
\begin{equation}
\begin{aligned}
\setlength\abovedisplayskip{3pt}
\setlength\belowdisplayskip{3pt}
\boldsymbol{m}_{e}^{(l)} & = \text{AGG}^{(l)}\left( \left\{ \boldsymbol{W}_{r}^{(l)}\boldsymbol{h}_{e'}^{(l-1)} \mid (e',r,e) \in \mathcal{T}_q \right\} \right), \\
\boldsymbol{h}_{e}^{(l)} & = \text{UPDATE}^{(l)}\left( \boldsymbol{h}_{e}^{(l-1)}, \boldsymbol{m}_{e}^{(l)} \right),
\end{aligned}
\label{eq_2}
\end{equation}
\noindent where $\boldsymbol{h}_e^{(l)}$ denotes the representation of node $e$ at the $l\text{-th}$ layer, and $\boldsymbol{W}_r^{(l)}$ is a learnable relation matrix. The functions $\text{AGG}^{(l)}$ and $\text{UPDATE}^{(l)}$ represent aggregation and update operations, respectively. Initial representations $\boldsymbol{h}_e^{(0)}$ and $\boldsymbol{W}_r^{(0)}$ are randomly initialized. After $L$ layers of message passing, $\boldsymbol{h}_e^{(L)}$ and $\boldsymbol{W}_r^{(L)}$ encode rich  structural information, which will be used for subsequent reasoning tasks. For brevity, we omit the superscripts and  denote them as $\boldsymbol{h}_e$ and $\boldsymbol{W}_r$, respectively.

\subsection{Path Prioritization Module}

Based on the query subgraph $\mathcal{G}_q$, we introduce a path prioritization module to identify multi-hop reasoning paths, which serve as logical chains of thought and support explainable reasoning. Unlike existing methods that indiscriminately retrieve all possible reasoning paths, often introducing substantial irrelevant or noisy data, our approach prioritizes more important reasoning paths, enabling more faithful and interpretable reasoning for LLMs.

\begin{figure}
    \centering
    \includegraphics[width=0.47\textwidth]{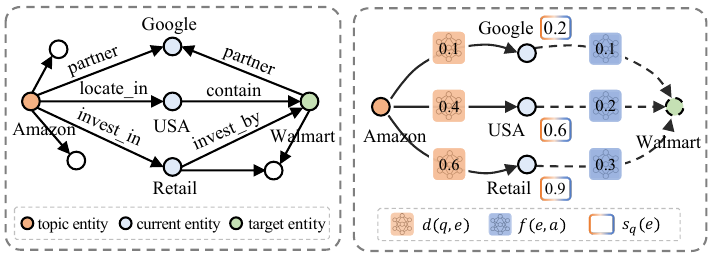}
    \caption{Illustration of the path priority function on the KG, where the current entity $e$ is evaluated based on the accumulative cost $d(q, e)$ and the estimated future cost $f(e, a)$.} \label{fig_astar}
\end{figure}

Inspired by the A* algorithm for path planning~\citep{zhu2023net, meng2024llm, zhuang2024toolchain}, we argue that \textit{an effective path prioritization mechanism should simultaneously consider the accumulative cost up to the current step and the estimated cost to reach the target}. Building on this insight, we design a novel path priority function to guide the search towards important reasoning paths. Specifically, it comprises two aspects: \textbf{the accumulative cost} $d(q, e)$, which measures the cost from the query $q$ to the current entity $e$, and \textbf{the estimated future cost} $f(e, a)$, which estimates the remaining cost to reach the target answer $a$. The overall priority score is defined as $s_q(e) = d(q, e) + f(e, a) $. Figure~\ref{fig_astar} illustrates this process on the KG.

However, designing an effective path priority function presents two key challenges. First, unlike the traditional A* algorithm typically applied to grid graphs, KGs are heterogeneous graphs where edges represent semantic relationships rather than geometric distances, making it non-trivial to define an appropriate cost function that measures the "semantic distances" between entities. Second, KGs are often large-scale with vast numbers of entities and relations. Consequently, directly applying the A* algorithm can easily lead to an intractably large search space, rendering the search inefficient or even infeasible.

To address these challenges, we propose a semantic-aware path priority function that efficiently discovers important reasoning paths. To be concrete, we define the accumulative cost $d(q,e)$ as the aggregation of the paths that connect the topic entities in the query $q$ to the current entity $e$. We compute its corresponding representation $\boldsymbol{d}(q,e)$, as follows:  
\begin{equation}
\begin{aligned}
\setlength\abovedisplayskip{3pt}
\setlength\belowdisplayskip{3pt}
\boldsymbol{d}(q,e) &= \sum_{\pi \in \Pi_{q \rightsquigarrow e}} \boldsymbol{d}(\pi) \\ &= \sum_{\pi \in \Pi_{q \rightsquigarrow e}} 
    \sum_{(e_{i-1}, r_{i}, e_{i}) \in \pi} \boldsymbol{w}_q(e_{i-1}, r_{i}, e_{i}),
\end{aligned}
\end{equation}

\noindent where $\Pi_{q \rightsquigarrow e}$ denotes the set of paths from the query $q$ to the entity $e$, and $\boldsymbol{w}_q(e_{i-1}, r_{i}, e_{i})= (\boldsymbol{h}_{e_{i-1}} \boldsymbol{W}_{r_{i}}  \boldsymbol{h}_{e_{i}})^\top\boldsymbol{q} $ denotes the semantic representation of the triple $(e_{i-1}, r_{i}, e_{i})$ in the path $\pi$, conditioned on the query $q$.

For the estimated future cost $f(e, a)$, since the target answer $a$ is unknown in advance, we reparameterize the answer $a$ using the topic entities and the query relation. Thus, we define the representation $\boldsymbol{f}(e, a)$ as:
\begin{equation}
\begin{aligned}
\boldsymbol{f}(e,a)&= \boldsymbol{f}([\boldsymbol{d}(q,e), \boldsymbol{q}]),
\end{aligned}
\end{equation}
\noindent where $\boldsymbol{f}(\cdot)$ is a feed-forward neural network, and $[\cdot, \cdot]$ concatenates two representations. Intuitively, the function $\boldsymbol{f}([\boldsymbol{d}(q, e), \boldsymbol{q}])$ compares the current representation $\boldsymbol{d}(q, e)$ with the query $\boldsymbol{q}$ to estimate the remaining cost. If $\boldsymbol{d}(q, e)$ is close to $\boldsymbol{q}$, the remaining cost will be close to 0, indicating that the entity $e$ is likely close to the correct answer. Finally, the overall priority score is computed as:
\begin{equation}
\begin{aligned}
s_q(e)&= \sigma\bigl( \text{MLP}\bigl( \boldsymbol{d}\left(q, e\right) + \boldsymbol{f}\left(e, a\right) \bigr) \bigr),
\end{aligned}
\end{equation}
\noindent where $\text{MLP}$ denotes a multi-layer perceptron, and $\sigma(\cdot)$ is the sigmoid function that normalizes the output to [0, 1]. In this way, we effectively measure semantic distances while significantly reducing the search space by restricting the retrieval scope to the query subgraph $\mathcal{G}_q$.

\subsubsection{Learning.} To learn the path priority function, we use the knowledge reasoning task as supervision, encouraging the model to assign higher priorities $s_q(e)$ to entities that are more important for reasoning. The loss is as follows:
\begin{equation}
\mathcal{L}=-\sum_{e \in \mathcal{A}_q} \log \left(s_q(e)\right)-\sum_{e \in \mathcal{G}_q \backslash \mathcal{A}_q} \log \left(1-s_q(e)\right),
\end{equation}
\noindent where $\mathcal{A}_q$ is the set of answers to the query $q$. Once the priority function $s_q(e)$ is learned, we repeatedly select the top-$K$ entities based on the learned priority scores during each iteration $t \in T$. As a result, we identify important reasoning paths $\Pi_q = \{\pi_i \mid \pi_i= e_{0}\xrightarrow{r_1} e_{1} \xrightarrow{r_2} \dots \xrightarrow{r_T} e_{T}\}_{i=1}^{N}$, where $N$ is the number of reasoning paths.

\subsection{Knowledge Reasoning Module}
Upon obtaining the important reasoning paths, we fine-tune LLMs via a two-phase training strategy: task-specific instruction tuning and path-wise preference alignment. This ensures that the model not only generates accurate responses but also develops logically consistent reasoning. In addition, it avoids the high overhead from multiple LLM calls.

\subsubsection{Task-Specific Instruction Tuning.} Initially, the instruction tuning phase is designed to improve the model's ability to interpret and solve reasoning problems by adhering to task-specific instructions. In this phase, we employ Supervised Fine-Tuning (SFT) to train the model on various KGR tasks. Specifically, the model takes the query $q$ and the important reasoning paths $\Pi_q$ as input, and generates the corresponding answers $\mathcal{A}_q$ as output. The SFT loss is formally defined as follows:
\begin{equation}
\mathcal{L}_{\text{SFT}}=-\mathbb{E}_{(q, \mathcal{A}_q) \sim \mathcal{D}_{\text{SFT}}} \big[\log P_{\phi}(\mathcal{A}_q \mid q, \Pi_q) \big ],
\end{equation}
\noindent where $\phi$ denotes the model parameters, and the task dataset is defined as $\mathcal{D}_{\text{SFT}}=\{q^{(i)}, \mathcal{A}_q^{(i)}\}_{i=1}^{N}$. The reasoning paths $\Pi_q$ are verbalized into textual prompts using an instruction template, as detailed in Appendix~\ref{appendix_template}. The model trained in this phase is denoted as $\mathcal{M}_{\text{sft}}$.

\subsubsection{Path-Wise Preference Alignment.} The second phase applies Direct Preference Optimization (DPO) to further align the model with more accurate and logically consistent reasoning based on preference paths. For each query $q$, we consider the input pairs labeled $(\Pi_q^w, \Pi_q^l)$, where $\Pi_q^w$ and $\Pi_q^l$ represent the preferred and less preferred paths, respectively. DPO refines the learned SFT model, $\mathcal{M}_{\text{sft}}$, by aligning it more closely with the preferred paths. The training objective for DPO is defined as follows:
\begin{equation}
\begin{aligned}
& \mathcal{L}_{\text{DPO}}\left(\mathcal{M} ; \mathcal{M}_{\text{sft}}\right)=-\mathbb{E}_{\left(q, \Pi_q^w, \Pi_q^l\right) \sim \mathcal{D}_{\text{DPO}}} \\
& {\left[\log \sigma\left(\beta \log \frac{\mathcal{M}\left(\Pi_q^w \mid q\right)}{\mathcal{M}\left(\Pi_q^l \mid q\right)}-\beta \log \frac{\mathcal{M}_{\text{sft}}\left(\Pi_q^w \mid q\right)}{\mathcal{M}_{\text{sft}}\left(\Pi_q^l \mid q\right)}\right)\right],}
\end{aligned}
\end{equation}
\noindent where $\beta$ is a parameter controlling the deviation from the base reference model $\mathcal{M}_{\text{sft}}$, and the comparison dataset is defined as $\mathcal{D}_{\text{DPO}}=\{q^{(i)}, \Pi_q^{w(i)}, \Pi_q^{l(i)}\}_{i=1}^{M}$. To construct these preference pairs ($\Pi_q^w$, $\Pi_q^l$), we treat the retrieved important reasoning paths $\Pi_q$ as the preferred paths $\Pi_q^w$, while the less preferred paths $\Pi_q^l$ are sampled from the remaining candidate paths in the subgraph $\mathcal{G}_q$. The instruction template for path alignment is provided in Appendix~\ref{appendix_template}.

\section{Experiments}
We conduct experiments to validate the effectiveness of our model and address key research questions. \textbf{RQ1:} How does PathMind perform compared to existing KGR methods? \textbf{RQ2:} What are the contributions of different modules to the overall performance? \textbf{RQ3:} How generalizable, scalable, and efficient is our PathMind framework? \textbf{RQ4:} How does PathMind perform interpretable reasoning?

\subsection{Experimental Settings}
\subsubsection{Datasets.} We evaluate PathMind on two popular benchmark datasets: WebQuestionSP (WebQSP)~\citep{yih2016value} and Complex WebQuestions (CWQ)~\citep{talmor2018web}. See~\citet{luo2024reasoning} for more details.

\subsubsection{Baselines.}
We compare our method against representative baselines, categorized into two groups: (1) \textit{Traditional KGR Methods}, including embedding-based (KVMem, NSM) and retrieval-based (GraftNet, ReaRev, SR+NSM) approaches; and (2) \textit{LLM-based KGR Methods}, including LLM reasoning (Qwen2-7B, Llama-2-7B, Llama-3.1-8B, GPT-4o), retrieval-augmented (BGE, MindMap, RoG, LightPROF, SubgraphRAG, GNN-RAG, GCR, EPERM), and synergy-augmented (KD-CoT, StructGPT, ToG, PoG, EffiQA, KnowPath) approaches. For fair comparison, we report results from the original papers. Additionally, we reproduce recent methods GNN-RAG and GCR using the same LLM to ensure consistent evaluation. Detailed descriptions of each baseline are provided in Appendix~\ref{appendix_baseline}.

\subsubsection{Evaluation Protocols.}
Following prior work, we evaluate model performance using two standard metrics: Hits@1 and F1. Hits@1 measures the proportion of instances where the top-1 predicted answer is correct, while F1 provides a balanced measure of answer coverage.

\subsubsection{Implementation Details.} 
In our implementation, we use Llama3.1-8B as the LLM backbone. For subgraph retrieval, we sample 3-hop neighborhoods to construct query subgraphs. In path prioritization, node and relation representations are learned using GNN, while query representations are encoded with the pre-trained BERT. We select the top-3 nodes at each iteration and set the maximum number of iterations to 2 for WebSQP and 4 for CWQ. For knowledge reasoning, we train the model for 3 epochs with a batch size of 2. The learning rate is set to 2e-5 with a warm-up ratio of 3e-2. In DPO training, the learning rate is 5e-6, and the hyperparameter $\beta$ is set to 0.1. The maximum input length for the LLM is set to 2048 tokens. Our model is implemented in PyTorch and trained on two NVIDIA A800 GPUs.

\subsection{Overall Comparison (RQ1)}
Table~\ref{tab_1_results} presents the main results on KGR using WebQSP and CWQ. (1) \textit{Overall Performance:} PathMind significantly outperforms all compared baselines, demonstrating superior reasoning capabilities over KGs. Specifically, our method improves Hits@1 by 0.8\% over the second-best method EPERM, on WebQSP. On the more challenging CWQ dataset, which involves multi-hop questions, PathMind achieves improvements of 5.1\% in Hits@1 and 3.9\% in F1 compared to the strong baseline GNN-RAG. These results highlight the effectiveness of our method, especially in handling complex reasoning tasks. (2) \textit{Analysis of Other Methods}: Interestingly, while GCR excels in Hits@1, its F1 score is relatively low, likely because the model tends to focus on the most probable answer. Among traditional methods, retrieval-based approaches outperform embedding-based methods. For example, SR+NSM, which uses path retrieval, achieves better performance, emphasizing the significance of reasoning paths. (3) \textit{LLM Performance and Challenges:} Despite the strong general capabilities of LLMs such as Llama-3.1-8B and GPT-4o, they still exhibit a significant performance gap compared to the best fine-tuned models. This underscores the inherent difficulty of answering complex queries using LLMs alone.

\begin{table}[ht]
\centering
\begin{tabular}{l@{\hspace{15pt}}cccc} 
\toprule
\multirow{2}{*}{\textbf{Methods}} & \multicolumn{2}{c}{\textbf{WebQSP}} & \multicolumn{2}{c}{\textbf{CWQ}} \\ 
\cmidrule(lr){2-3} \cmidrule(lr){4-5}
& \textbf{Hits@1} & \textbf{F1} & \textbf{Hits@1} & \textbf{F1} \\ 
\midrule
KV-Mem         & 0.467 & 0.345 & 0.184 & 0.157 \\
GraftNet       & 0.664 & 0.604 & 0.368 & 0.327 \\
NSM            & 0.687 & 0.628 & 0.476 & 0.424 \\
SR+NSM         & 0.689 & 0.641 & 0.502 & 0.471 \\
ReaRev         & 0.764 & 0.709 & 0.529 & 0.478 \\ 
\midrule
Qwen2-7B       & 0.508 & 0.355 & 0.253 & 0.216 \\
Llama-2-7B     & 0.564 & 0.365 & 0.284 & 0.214 \\ 
Llama-3.1-8B   & 0.555 & 0.348 & 0.281 & 0.224 \\ 
GPT-4o         & 0.618 & 0.436 & 0.382 & 0.329 \\
\midrule
KD-CoT         & 0.686 & 0.525 & 0.557 & --    \\
StructGPT      & 0.726 & 0.637 & 0.543 & 0.496 \\
MindMap        & 0.649 & 0.471 & 0.488 & 0.433 \\
ToG            & 0.826 & --    & 0.685 & --    \\
RoG            & 0.857 & 0.708 & 0.626 & 0.562 \\
LightPROF      & 0.838 & --    & 0.593 & --    \\
KnowPath       & 0.841 & --    & 0.679 & --    \\
GNN-RAG$^*$    & 0.864 & 0.690 & 0.673 & \underline{0.591} \\
SubgraphRAG    & 0.866 & 0.706 & 0.472 & 0.570 \\
EPERM          & \underline{0.888} & \underline{0.724} & 0.662 & 0.589 \\ 
GCR$^*$           & 0.883   & 0.654    & \underline{0.686}    & 0.532  \\ 			
\midrule
PathMind (ours) & \textbf{0.895}    & \textbf{0.728}    & \textbf{0.707}     & \textbf{0.614}     \\
\bottomrule
\end{tabular}
\caption{Performance comparison of PathMind with baselines on two datasets. (\textbf{bold} denotes the best results, \underline{underline} denotes the second best results, and "*" marks results reproduced using Llama3.1-8B for fair comparison.). }
\label{tab_1_results}
\end{table}

\subsection{Ablation Study (RQ2)}
\subsubsection{Performance on Model Ablation.} To evaluate the effect of different components of PathMind, we compare three variants: (1) \textit{w/o Prioritization}, removing the path prioritization module; (2) \textit{w/o Alignment},  excluding path-wise preference alignment; and (3) \textit{w/o Training},  without the two-phase training strategy. As shown in Table~\ref{tab_ablation}, removing path prioritization makes the model rely solely on the query subgraph, resulting in a significant decline in performance on WebQSP and CWQ. This finding validates the critical role of identifying and selecting explicit reasoning paths. The absence of alignment (via DPO) yields inferior results, indicating the necessity of aligning the model with preferred reasoning paths.  Furthermore, without training, the model struggles to interpret the structural knowledge of KGs, leading to substantial performance degradation.

\begin{table}[ht]
\centering

\begin{tabular}{l@{\hspace{8pt}}cccc}
\toprule
\multirow{2}{*}{\textbf{Variants}} & \multicolumn{2}{c}{\textbf{WebQSP}} & \multicolumn{2}{c}{\textbf{CWQ}} \\
\cmidrule(lr){2-3} \cmidrule(lr){4-5}
& \textbf{Hits@1} & \textbf{F1} & \textbf{Hits@1} & \textbf{F1} \\
\midrule
PathMind & 0.895 & 0.728 & 0.707 & 0.614  \\
\; w/o Priorization & 0.840 & 0.662 & 0.643 & 0.561  \\
\; w/o Alignment & 0.871 & 0.695 & 0.672 & 0.586 \\
\; w/o Training & 0.668 & 0.480 & 0.413 & 0.274 \\
\bottomrule
\end{tabular}
\caption{Model ablation study of our PathMind framework.}
\label{tab_ablation}
\end{table}

\begin{table}[ht]
\centering
\begin{tabular}{l@{\hspace{10pt}}cccc}
\toprule
\multirow{2}{*}{\textbf{Strategies}} & \multicolumn{2}{c}{\textbf{WebQSP}} & \multicolumn{2}{c}{\textbf{CWQ}} \\
\cmidrule(lr){2-3} \cmidrule(lr){4-5}
& \textbf{Hits@1} & \textbf{F1} & \textbf{Hits@1} & \textbf{F1} \\
\midrule
Random Paths    & 0.356  & 0.104  & 0.268  & 0.079  \\
Shortest Paths  & 0.854  & 0.681  & 0.662  & 0.578 \\
Important Paths & 0.895 & 0.728 & 0.707 & 0.614   \\
\bottomrule
\end{tabular}
\caption{Impact on different path prioritization strategies.}
\label{tab_path_strategy}
\end{table}

\begin{figure}[ht]
    \centering
    \includegraphics[width=0.48\textwidth]{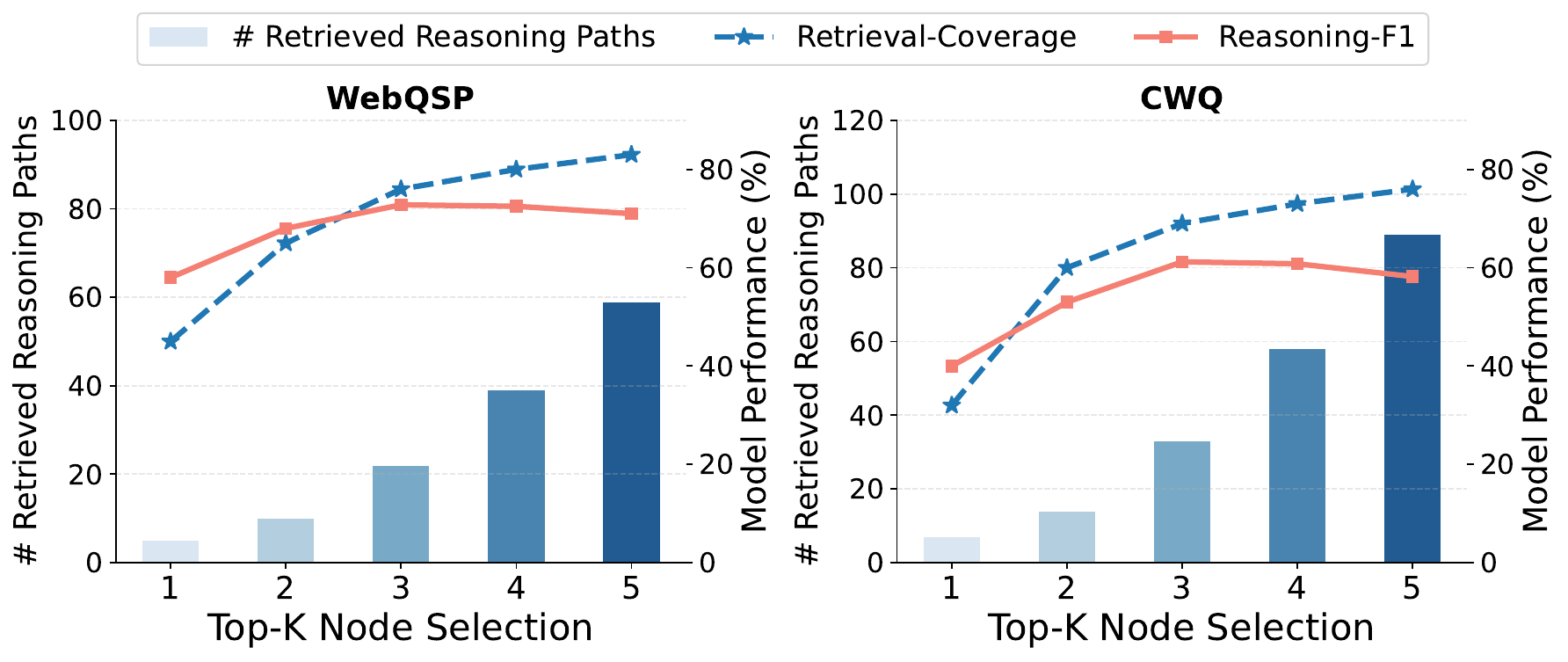}
    \caption{Effect on varying numbers of node selection.} \label{fig_ablation_node}
\end{figure}

\subsubsection{Performance on Path Prioritization.}

To further investigate the necessity of path prioritization, we compare three path selection strategies: \textit{random paths}, \textit{shortest paths}, and \textit{important paths (ours)}. Random paths are constructed via random walks, while shortest paths are obtained using the GNN-RAG method. As shown in Table~\ref{tab_path_strategy}, our method significantly outperforms the other two strategies, demonstrating the effectiveness of prioritizing semantically relevant paths. Although other strategies enable efficient traversal, they often introduce irrelevant noise and hinder accurate reasoning. In addition, we find that path selection has a more pronounced impact on CWQ compared to WebQSP. We hypothesize that this is due to the inherently multi-hop nature of complex questions in CWQ, where effective path selection is crucial for supporting logical reasoning. More variations on the path priority function are provided in Appendix~\ref{appendix_experiment}.

\subsubsection{Performance on Hyperparameter Setting.} 

We consider two key hyperparameters: $T$, the maximum number of iterations, and $K$, the maximum number of nodes selected per iteration. Based on the characteristics of each dataset, we set $T=2$ for WebQSP and $T=4$ for CWQ. To assess the impact of varying $K$ on model performance, we conduct a series of experiments. As depicted in Figure~\ref{fig_ablation_node}, performance improves with increasing $K$.  However, when $K$ exceeds 3, the F1 scores on both datasets tend to decline, likely due to the inclusion of irrelevant entities that obscure critical information. Consequently, we select $K=3$ as the optimal setting to identify important reasoning paths.

\begin{figure*}
    \centering
    \includegraphics[width=0.99\textwidth]{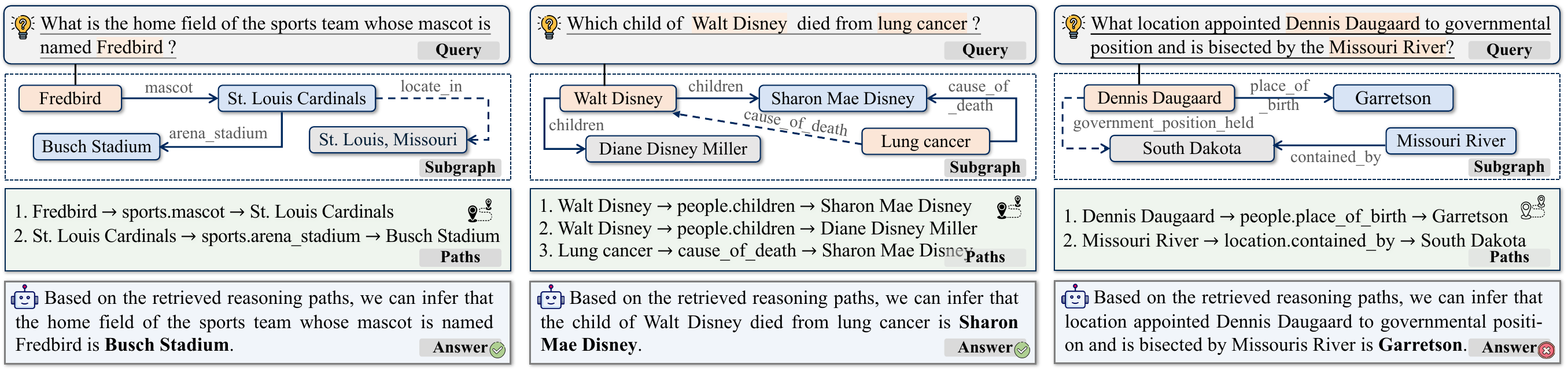}
	\caption{Illustrations of interpretable reasoning of PathMind on CWQ (solid lines indicate retrieved important paths).} \label{fig_casestudy}
\end{figure*}

\subsection{Further Analysis (RQ3)}

\subsubsection{Generalizability Across LLMs.}
To assess the generalizability of PathMind across various LLMs, we conduct experiments using representative models, including Qwen2-7B, Llama2-7B, and Llama3.1-8B. From the results in Table \ref{tab_llm}, we can see that PathMind achieves outstanding performance across these LLMs, underscoring the effectiveness and adaptability of our proposed framework. In particular, when fine-tuned on Llama3.1-8B, our method attains higher accuracy than other LLMs, suggesting that more advanced models can significantly boost overall performance.

\begin{table}[ht]
\centering
\begin{tabular}{lcccc}
\toprule
\multirow{2}{*}{\textbf{Backbones}} & \multicolumn{2}{c}{\textbf{WebQSP}} & \multicolumn{2}{c}{\textbf{CWQ}} \\
\cmidrule(lr){2-3} \cmidrule(lr){4-5}
& \textbf{Hits@1} & \textbf{F1} & \textbf{Hits@1} & \textbf{F1} \\
\midrule
Llama2-7B\quad\quad & 0.864 & 0.687 & 0.652 & 0.573  \\
Qwen2-7B & 0.872 & 0.693 & 0.665 & 0.580 \\
Llama3.1-8B & 0.895 & 0.728 & 0.707 & 0.614  \\
\bottomrule
\end{tabular}
\caption{Transferability of PathMind across various LLMs.}
\label{tab_llm}
\end{table}

\begin{figure}[ht]
    \centering
    \begin{subfigure}[b]{0.48\linewidth}
        \centering
        \includegraphics[width=\linewidth]{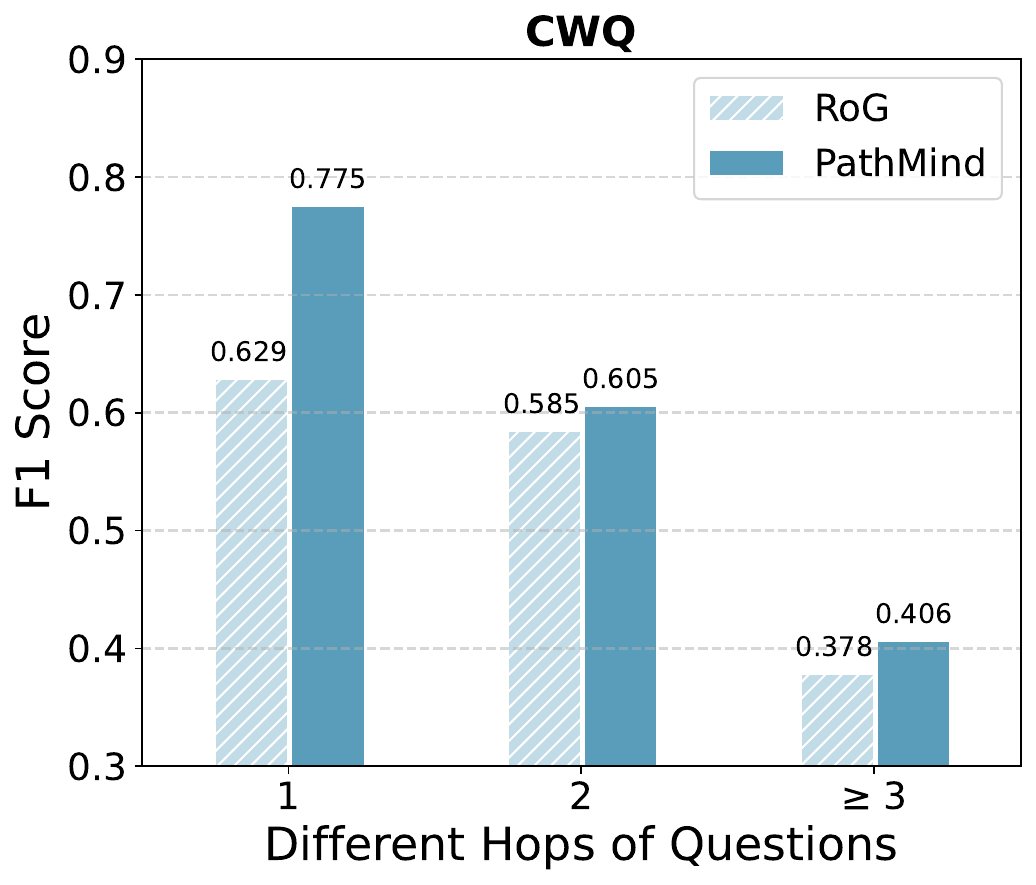}
    \end{subfigure}
    \hfill
    \begin{subfigure}[b]{0.48\linewidth}
        \centering
        \includegraphics[width=\linewidth]{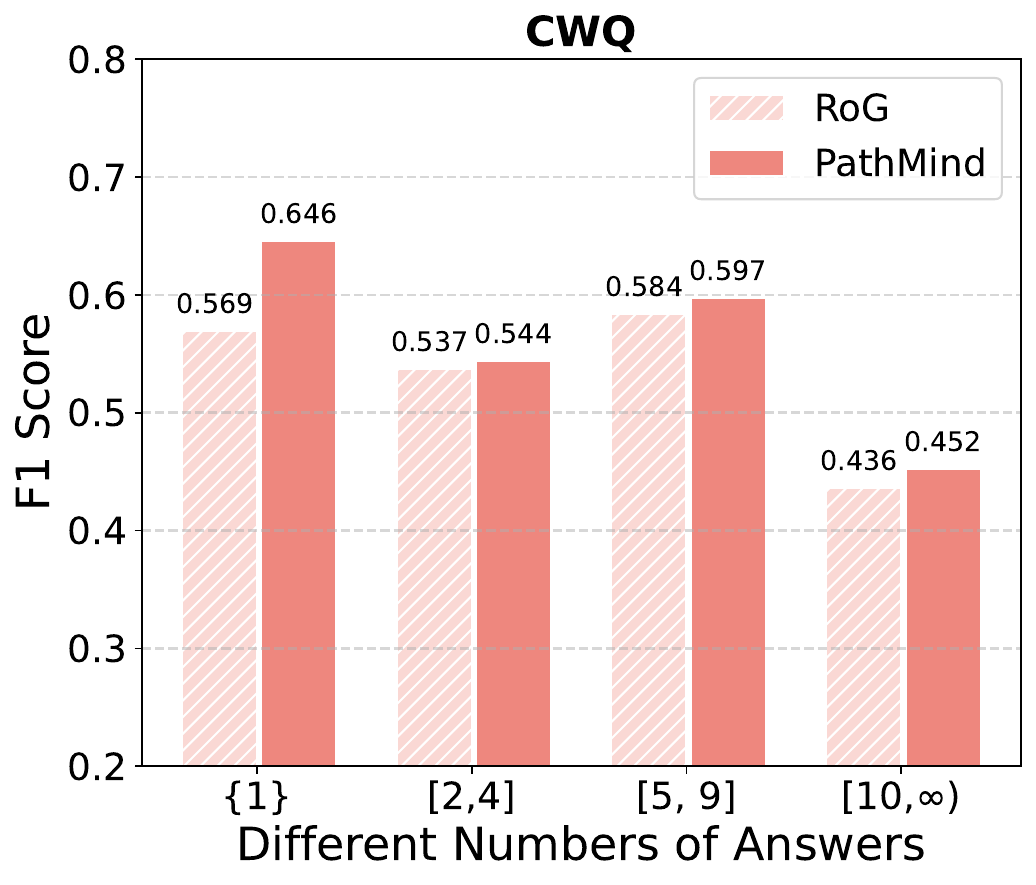}
    \end{subfigure}
    \caption{Scalability of PathMind with different question hops and different answer numbers on CWQ.}
    \label{tab_answer_result}
\end{figure}

\subsubsection{Reasoning Scalability Analysis.}
To better understand the scalability of our model, we analyze its performance across reasoning hops and answer numbers. As shown in Figure~\ref{tab_answer_result}, PathMind consistently surpasses RoG, demonstrating strong robustness and scalability. In simple scenarios, our method effectively identifies important paths, yielding substantial improvements. As task complexity increases, PathMind maintains its leading performance, thanks to its ability to disregard numerous irrelevant paths, thereby reducing interference from extraneous information. 

\subsubsection{Model Efficiency Evaluation.}
To evaluate model efficiency, we analyze the average runtime, number of LLM calls, and number of input tokens in Table~\ref{tab_efficiency}. Specifically, synergy-augmented methods, such as PoG, achieve superior results but require more LLM calls and input tokens due to their iterative reasoning process. In contrast, retrieval-augmented methods benefit from efficient subgraph retrieval, which significantly reduces runtime and computational costs. Our method strikes a favorable balance between performance and efficiency, particularly by reducing input tokens through the identification of important paths.

\begin{table}[ht]
\centering
\begin{tabular}{l@{\hspace{10pt}}cccc}
\toprule
\textbf{Methods} & \textbf{Hit (\%)} & \textbf{Time (s)} & \textbf{\# Calls} & \textbf{\# Tokens} \\
\midrule
\multicolumn{5}{c}{\textit{Synergy-Augmented Methods}} \\
ToG            & 0.751  & 16.14  & 11.6  & 7,069   \\
EffiQA         & 0.829  & --     & 7.3   & --      \\
PoG            & 0.873  & 16.80  & 9.0   & 5,518   \\
\midrule
\multicolumn{5}{c}{\textit{Retrieval-Augmented Methods}} \\
RoG            & 0.857  & 2.60   & 2     & 521     \\
GNN-RAG        & 0.864  & \textbf{1.52}   & \textbf{1}     & 414     \\
GCR        & 0.883  & 3.60   & 2     & 231     \\ 
\midrule
PathMind       & \textbf{0.895}    & 2.23     & \textbf{1}     & \textbf{216} \\
\bottomrule
\end{tabular}
\caption{Efficiency comparison of PathMind on WebQSP.}
\label{tab_efficiency}
\end{table}

\subsection{Case Study (RQ4)}

We further present three complex cases in CWQ to demonstrate the interpretable reasoning process of PathMind. As illustrated in Figure~\ref{fig_casestudy}, in Case 1, the method correctly extracts the two-hop reasoning path {\small $Fredbird \xrightarrow{mascot} St.\ Louis\ Cardinals \xrightarrow{arena\ stadium} Busch\ Stadium$}, which is consistent with human cognition. In Case 2, although the retrieved reasoning paths contain noises, such as "the child of Walt Disney" corresponding to two facts: {\small $Walt\, Disney \xrightarrow{children} Sharon\, Mae\, Disney $} and {\small $Walt\, Disney \xrightarrow{children} Diane\, Disney\, Miller$}, the reasoning module can successfully discern valuable evidence and perform faithful reasoning. In Case 3, our model makes an incorrect prediction due to the missing important path {\small $Dennis\ Daugaard \xrightarrow{gov.\, pos.\, held} South\ Dakota $}, which highlights the significance of path  identification from KGs to augment the reasoning capabilities of LLMs.

\section{Conclusion}

In this paper, we introduce PathMind, a novel framework designed to enhance LLMs for KGR. The core idea is to guide LLMs using important reasoning paths, enabling more faithful and interpretable reasoning. Our method comprises three key components: subgraph retrieval, path prioritization, and knowledge reasoning.  Extensive results and model analysis demonstrate that PathMind consistently outperforms competitive baselines across various KGR benchmarks. 



\bibliography{main}

\appendix

\section{Instruction Templates for Reasoning}\label{appendix_template}
In the knowledge reasoning module, we design two instruction templates. (1) For task-specific instruction tuning, we design a instruction template that verbalizes the retrieved reasoning paths into textual prompts. (2) For path-wise preference alignment, we craft instruction that guides LLMs to align with the preferred reasoning paths.

\begin{figure}[ht]
\centering
\begin{tcolorbox}[ colback=white,colframe=black,left=3pt,right=3pt,
title=Instruction templates for task-specific instruction tuning, fonttitle=\footnotesize]
\footnotesize
\textbf{Instruction:} Based on the reasoning paths, please answer the given question. Please keep the answer as simple as possible and return all the possible answers as a list.

\bigskip
\textbf{Reasoning Paths:}~ \textless Reasoning Paths\textgreater \\
\textbf{Question:}~\textless Question\textgreater \\
\textbf{Answer:}~\textless Answer\textgreater
\end{tcolorbox}
\end{figure}

\begin{figure}[ht]
\centering
\begin{tcolorbox}[ colback=white,colframe=black,left=3pt,right=3pt,
title=Instruction templates for path-wise preference alignment, fonttitle=\footnotesize]
\footnotesize
\textbf{Instruction:}~Given the question, please generate coherent reasoning paths that can support answering it.

\bigskip
\textbf{Question:}~ \textless Question\textgreater \\
\textbf{Preferred Paths:}~\textless Preferred Paths \textgreater \\
\textbf{Less Preferred Paths:}~\textless Less Preferred Paths \textgreater

\end{tcolorbox}
\end{figure}

\section{Detailed Descriptions of Baselines}\label{appendix_baseline}
For a comprehensive comparison, we provide the detailed descriptions of the compared baselines.

\begin{itemize}[leftmargin=*, itemsep=0pt]
\item \textbf{Traditional KGR methods.} 
\textit{(i) Embedding-based methods:} 
KVMem~\citep{miller2016key} and NSM~\citep{he2021improving}
\textit{(ii) Retrieval-based methods:} 
GraftNet~\citep{sun2018open},
SR+NSM~\citep{zhang2022subgraph},
ReaRev~\citep{mavromatis2022rearev}

\item \textbf{LLM-based KGR methods.} 
\textit{(i) Retrieval-Augmented methods:}
BGE~\citep{zhang2023retrieve},
MindMap~\citep{wen2024mindmap},
RoG~\citep{luo2024reasoning},
GNN-RAG~\citep{mavromatis2025gnn},
SubgraphRAG~\citep{li2025simple},
LightPROF~\citep{ao2025lightprof},
EPERM~\citep{long2025eperm},
GCR~\citep{luo2025graph}
\textit{(ii) Synergy-Augmented methods:} 
KD-CoT~\citep{wang2023knowledge},
EffiQA~\citep{dong2025effiqa},
StructGPT~\citep{jiang2023structgpt},
ToG~\citep{sun2024think},
PoG~\citep{chen2024plan},
KnowPath~\citep{zhao2025knowpath},
\end{itemize}

\section{Additional Experiment Results}\label{appendix_experiment}

\subsubsection{Performance on Path Priority Function.}
For the path priority function, we analyze its two key components: the accumulative cost $d(q,e)$ and the estimated future cost $f(e, a)$. As shown in Table~\ref{tab_path_function}, the accumulative cost plays a more critical role since it reflects the actual cost incurred along the path. The best performance occurs when both costs are combined, providing a more comprehensive evaluation.
\begin{table}[ht]
\centering
\begin{tabular}{l@{\hspace{14pt}}cccc}
\toprule
\multirow{2}{*}{\textbf{Variants}} & \multicolumn{2}{c}{\textbf{WebQSP}} & \multicolumn{2}{c}{\textbf{CWQ}} \\
\cmidrule(lr){2-3} \cmidrule(lr){4-5}
& \textbf{Hits@1} & \textbf{F1} & \textbf{Hits@1} & \textbf{F1} \\
\midrule
Accum-Cost    & 0.878  & 0.714  & 0.683  & 0.602  \\
Future-Cost  & 0.831  & 0.672  & 0.635  & 0.576 \\
Full Costs & 0.895 & 0.728 & 0.707 & 0.614   \\
\bottomrule
\end{tabular}
\caption{Impact on different path prioritization strategies.}
\label{tab_path_function}
\end{table}

\subsubsection{Supplementary Scalability Study.}
\begin{figure}[!htbp]
    \centering
    \begin{subfigure}[b]{0.48\linewidth}
        \centering
        \includegraphics[width=\linewidth]{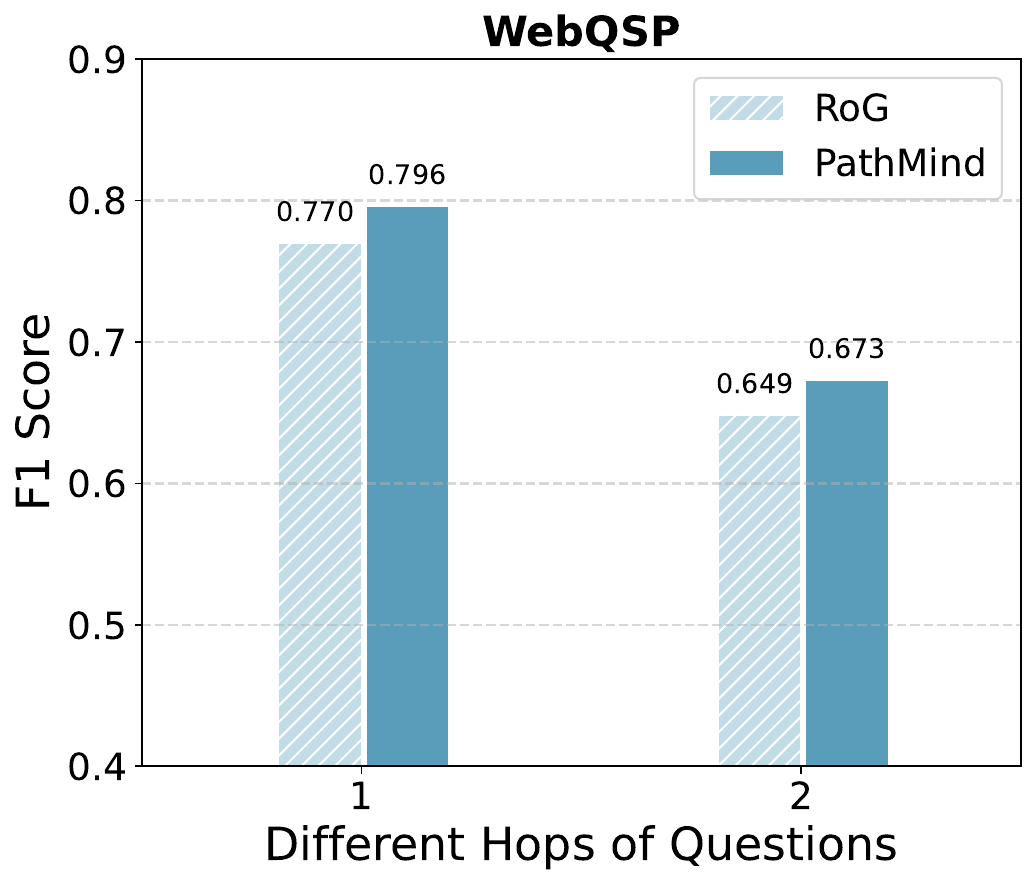}
    \end{subfigure}
    \hfill
    \begin{subfigure}[b]{0.48\linewidth}
        \centering
        \includegraphics[width=\linewidth]{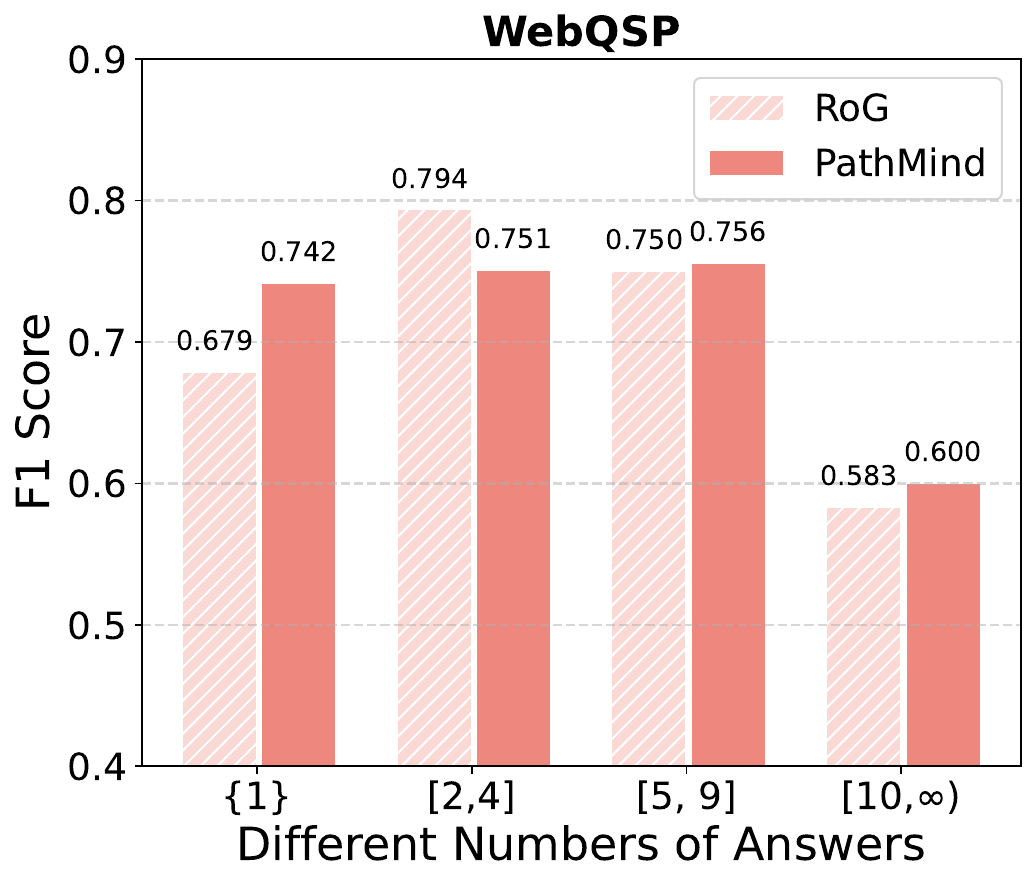}
    \end{subfigure}
    \caption{Scalability of PathMind with different question hops and different answer numbers on WebQSP.}
    \label{tab_scalability_webqsp}
\end{figure}

The statistics of reasoning hops and the answer numbers are presented in Table~\ref{tab_question_hop} and Table~\ref{tab_answer_number}, respectively. From the results, we can see that: (1) WebQSP is dominated by simple (single-hop) questions, whereas CWQ contains more complex questions that often require two or more reasoning hops; and (2) multi-answer questions ($\geq$2 answers) are more prevalent in WebQSP, while CWQ tends to favor single-answer cases. Model performance across different numbers of reasoning hops and answers on WebQSP is illustrated in Figure~\ref{tab_scalability_webqsp}.

\begin{table}[ht]
\centering
\begin{tabular}{l@{\hspace{12pt}}c@{\hspace{12pt}}c@{\hspace{12pt}}c@{\hspace{12pt}}c@{\hspace{12pt}}}
\toprule
\textbf{Dataset} & 1-hop & 2-hop & $\geq$ 3-hop \\
\midrule
WebQSP & 65.5\% & 34.5\%   & 0.0\% \\
CWQ    & 40.9\% & 38.3\%   & 20.8\%  \\
\bottomrule
\end{tabular}
\caption{Reasoning hops on WebQSP and CWQ.}
\label{tab_question_hop}
\end{table}

\begin{table}[ht]
\centering
\begin{tabular}{lc@{\hspace{3pt}}c@{\hspace{3pt}}c@{\hspace{3pt}}c@{\hspace{3pt}}}
\toprule
\textbf{Dataset} & \#Ans=1 & 2$\geq$\#Ans$\leq$4 & 5$\geq$\#Ans$\leq$9 & \#Ans$\geq$10 \\
\midrule
WebQSP & 51.2\% & 27.4\% & 8.3\%  & 12.1\% \\
CWQ    & 70.6\% & 19.4\% & 6.0\%  & 4.0\%  \\
\bottomrule
\end{tabular}
\caption{Answer numbers on WebQSP and CWQ.}
\label{tab_answer_number}
\end{table}

\end{document}